\documentclass[conference]{IEEEtran}
\IEEEoverridecommandlockouts

\usepackage{cite}
\usepackage{amsmath,amssymb,amsfonts}
\usepackage{algorithmic}
\usepackage{graphicx}
\usepackage{textcomp}
\usepackage{xcolor}

\usepackage{comment}
\usepackage{cleveref}
\usepackage[table]{xcolor}
\usepackage{booktabs}
\usepackage{url}

\def\BibTeX{{\rm B\kern-.05em{\sc i\kern-.025em b}\kern-.08em
    T\kern-.1667em\lower.7ex\hbox{E}\kern-.125emX}}
\begin{document}

\title{From Monolingual to Multilingual: Evaluating Mamba for ASR in South African Languages
}
\author{\IEEEauthorblockN{Jesujoba O. Alabi$^1$, Julian Herreilers$^2$,  Badr M. Abdullah$^1$, Dietrich Klakow$^{1}$} 
\IEEEauthorblockA{\textit{$^1$Saarland University, Germany, $^2$Saigen, South Africa}  \\
}
}

\maketitle

\begin{abstract}
Recent advances in automatic speech recognition (ASR) have explored different sequence models, including Conformer-based models and newer state space models such as Mamba. Although prior work has evaluated these architectures in multiple languages, their effectiveness in African languages remains underexplored. In this work, we evaluate Mamba for ASR on seven South African languages. In monolingual experiments, each model is trained on 50 hours of speech per language, and we compare Mamba to a Conformer baseline of similar parameter scale. Mamba achieves similar recognition accuracy to Conformer while using fewer computational resources and training faster. We further evaluate generalization in this setting and find that both models struggle to generalize to speech that is much longer than what they were trained on. We then study multilingual ASR using Mamba models, where the baseline is pooling all languages together. On top of this, we tested three extensions: training with language-family information by adding both language and language-family embeddings as biases to the downsampled acoustic representations, and multitask learning with a CTC ASR objective and a language identification (LID) head. We find that multilingual training consistently improves performance over monolingual training. However, adding explicit language information does not improve in-domain performance but does improve cross-corpus robustness. We conducted ablation studies in low-resource multilingual settings using 5-hour and 10-hour per-language training data, where we observed gains from using language embeddings and further demonstrated that removing or altering them hurt model performance. Lastly, we analysed these embeddings and find that they do not capture linguistic similarity in a typological sense, but instead act as task-specific control vectors.

\end{abstract}


\definecolor{OIblue}{HTML}{0072B2}
\definecolor{OIorange}{HTML}{E69F00}
\definecolor{OIgreen}{HTML}{009E73}
\definecolor{OIvermillion}{HTML}{D55E00}
\definecolor{OIpurple}{HTML}{CC79A7}
\definecolor{LightAsh}{HTML}{E6E6E6}

\section{Introduction}

Automatic speech recognition (ASR) research has advanced significantly with the development of powerful sequence modeling architectures~\cite{radford2023robust,kim2023branchformer,prabhavalkar2023end}. Attention-based models such as Conformer~\cite{gulati20_interspeech} have become strong baselines due to their ability to capture both local and global dependencies through the integration of self-attention and convolutional modules. More recently, state space model (SSM)-based architectures such as Mamba~\cite{gu2024mamba} have emerged as efficient alternatives, offering linear-time sequence modeling and favorable memory scaling compared to quadratic-time attention mechanisms, making them particularly attractive for long-form speech processing.

Despite these advances, research on Mamba for ASR has primarily focused on high-resource languages, a small number of low-resource languages, long-context ASR, multilingual ASR, and streaming ASR, typically in comparison with Conformer-based and other architectures~\cite{zevallos25_interspeech,moriya25_interspeech,Ali2025MLMATM}. Relatively little attention has been paid to African languages, which face a  ``low-resource double bind"~\cite{ahia-etal-2021-low-resource}: limited annotated speech data and constrained computational resources. Although recent efforts have expanded speech resources for African languages~\cite{marivate2025swivuriso0,Diack2026WAXALAL}, it remains necessary to understand how Mamba performs under conditions of limited data and long-form speech recognition.

Long-form ASR is important in low-resource settings, where real-world speech exhibits substantial variation in utterance length, ranging from short prompted phrases to extended spontaneous speech. However, limited data availability often results in insufficient coverage of long-utterance structures during training, making it difficult for models to learn robust representations across sequence lengths. Prior work on long-form ASR in high-resource languages, including the study by \cite{Flynn2026BeyondTU}, has shown that modeling long utterances introduces challenges in maintaining robustness over over long inputs and handling distributional shifts in sequence length. These challenges are likely to be amplified in low-resource settings. This highlights the difficulty of building robust ASR systems under limited supervision and motivates the use of multilingual training strategies, which are commonly employed to mitigate data scarcity in low-resource ASR.

To address these gaps, we conduct a systematic study of Mamba for ASR across seven South African languages. We first compare Mamba against a Conformer-based encoder baseline in monolingual settings, where each model is trained on approximately 50 hours of transcribed speech per language. Our analysis focuses on long-context recognition behavior and length generalization. We then extend the study to the multilingual setting using Mamba-based models. In this setting, we consider a pooled multilingual baseline and three extensions for incorporating language information: (i) adding language embeddings as biases to the acoustic representations, (ii) extending (i) by also incorporating language-family embeddings as biases, and (iii) a multitask learning setup with a connectionist temporal classification (CTC) ASR objective and an auxiliary language identification (LID) head. We further analyze approach (ii) under low-resource multilingual conditions (5-hour and 10-hour per-language settings), a setup commonly used to mitigate data scarcity in ASR, in order to study the effect of explicit language conditioning. Additionally, we investigate the representations learned by the language embeddings.

Our results show performance differences across languages in the monolingual setting, with Nguni languages being more challenging than Sotho languages. Mamba achieves performance comparable to Conformer models of similar parameter size, while using fewer computational resources and training faster, indicating that SSMs can match strong attention-based baselines for South African ASR. Both models show limited robustness to speech that is substantially longer than what they were trained on, highlighting challenges in length generalization. In the multilingual setting, joint training consistently improves performance over monolingual models, confirming the benefits of cross-lingual parameter sharing. However, incorporating explicit language or language-family information does not improve in-domain performance but improves cross-corpus robustness. Furthermore, in low-resource multilingual settings, explicit language embeddings provide consistent performance gains. Further analysis shows that these embeddings do not capture linguistic similarity in a typological sense, but instead act as task-specific steering vectors.

\begin{table}
    \centering
    \caption{Dataset statistics for our experiments. Only the test splits of NCHLT and FLEURS were used.}
    \vspace{-8pt} 
    \scalebox{0.83}{
    \begin{tabular}{l|ccccccc}
    \toprule
    \textbf{Bucket} &	\textbf{nbl}	& \textbf{sot} &	\textbf{tsn} &	\textbf{tso}	& \textbf{ven}	& \textbf{xho} & \textbf{zul} \\
    \midrule
    \multicolumn{8}{l}{\textbf{\textit{Swivuriso (train)}}}\\
    Dur (h)	& 50.0 & 50.0 & 50.0 & 50.0 & 50.0 & 50.0 & 50.0 \\
    Utt. &	13.6k &	11.6k	& 14.5k &	11.6k	& 11.8k	& 11.4k	& 9.3k \\
    \multicolumn{8}{l}{\textbf{\textit{Swivuriso: (dev test)}}}\\
    Dur (h)	& 14.2 & 25.1 & 25.8 & 25.7 & 11.5 & 25.6 & 25.7 \\
    Short Utt. & 2266	&2717&	4304	&3500	&1195	&2470 &	1810 \\
    Long Utt. &	424	& 1002 &	965 &	894	& 460 &	1169	& 1215 \\
         \midrule
    \multicolumn{8}{l}{\textbf{\textit{NCHLT (test)}}}\\
      Utt.   & 3108 & 2722 & 2889 & 2905 & 2805 & 2770 & 2802 \\
    \multicolumn{8}{l}{\textbf{\textit{FLEURS (test)}}}\\
       Utt.  & -  &  & - & - & -  & 1041 & 854 \\
    \bottomrule
    \end{tabular}
    }
    \label{tab:data_stat}
    \vspace{-8pt} 
\end{table}

\section{South African Language ASR}

\textbf{Language Characteristics:} 
South Africa is a multilingual country with 11 official languages, as recognized in the Constitution, with South African Sign Language also recently granted official status. Most of these languages belong to the Bantu language family, while Afrikaans and English are Germanic Indo-European languages. 
The Bantu languages spoken in South Africa include the Nguni languages (isiZulu, isiXhosa, siSwati, and isiNdebele), which together with Xitsonga (from the Tswa-Ronga branch) form the broader Nguni-Tsonga grouping in many linguistic classifications. They also include the Sotho-Tswana languages (Sesotho, Sesotho sa Leboa, and Setswana), as well as Tshivenda, which belongs to the Venda branch and is often classified within a broader Sotho-Venda (or Sotho-Makua-Venda) grouping.
All these languages use Latin scripts, and code-switching with English or other languages is common in informal speech. 

\textbf{ASR Resources:} A major challenge for ASR research in African languages is resource scarcity, particularly the limited availability of large, high-quality training and evaluation datasets~\cite{alabi-etal-2025-charting,imam-etal-2025-automatic}. In the case of South African languages, several resources exist, including Codeswitch Soap opera~\cite{van-der-westhuizen-niesler-2018-first}, NCHLT Speech corpus~\cite{barnard14_sltu,BADENHORST2022107860}, Lwazi ASR~\cite{gumede-plauche-2009-initial}, Common Voice~\cite{ardila-etal-2020-common}, FLEURS~\cite{conneau-etal-2020-fleurs}, Vuk’uzenzele (ViXSD)~\cite{rajab-etal-2025-esethu}, and the recently released Swivuriso~\cite{marivate2025swivuriso0} dataset, which covers seven languages, while NCHLT Speech and Lwazi ASR cover ten. These datasets provide a foundation for ASR development.

\textbf{ASR Modeling:} 
Given the limited resources for African languages, several efforts have focused on developing ASR systems for South African languages. In particular, prior work has addressed code-switched speech involving English, including comparisons of bilingual and multilingual systems and acoustic modeling tailored to these linguistic contexts, with an emphasis on languages such as isiZulu, isiXhosa, Setswana, and Sesotho~\cite{ylmaz18c_interspeech,BISWAS2022101262,biswas19b_interspeech,biswas-etal-2020-semi-supervised}. Furthermore, multilingual speech encoders have been developed to improve representations for these languages~\cite{boito2024mhubert,alabi25_interspeech}. Despite this progress, there remains a need for resource-efficient ASR models, such as Mamba-based models, that maintain robust performance under limited training data, computational constraints, and varying utterance lengths.



\section{Experimental Setup}
\subsection{Datasets}

For our experiments and analysis, we use ASR data from three sources, which are:

\textbf{Swivuriso:}~\cite{marivate2025swivuriso0} is a large speech corpus for South African languages that covers seven languages, including isiNdebele (\texttt{nbl}), Sesotho (\texttt{sot}), Setswana (\texttt{tsn}), Xitsonga (\texttt{tso}), Tshivenda (\texttt{ven}), isiXhosa (\texttt{xho}), and isiZulu (\texttt{zul}). It contains scripted and unscripted speech across domains such as agriculture, healthcare, and general conversation, with over 90 speakers per language. The transcriptions include diacritics, punctuations, special symbols such as \texttt{[cs]} to denote code-switching, \texttt{[pause]} for pause, and \texttt{[?]}.

For this study, we define \textbf{short utterances} as speech segments between 0 and 30 seconds, reflecting typical ASR training conditions. Segments longer than this are considered \textbf{long utterances}. We sample 50 hours of short utterances per language to construct a balanced dataset comprising scripted and unscripted speech from the \emph{train} set. Short utterances from the \emph{dev} set are used for validation, while the \emph{dev\_test} set is split into short and long subsets for evaluation. \Cref{tab:data_stat} summarizes the statistics of the data set. In the test split, short utterances outnumber long utterances by at least a factor of two in most languages, except isiZulu.

\textbf{NCHLT Speech corpus:}~\cite{barnard14_sltu,BADENHORST2022107860} is another large speech corpus that covers all 11 official South African languages. It contains more than 100 hours of scripted speech and more than 200 speakers per language. We used only the test splits of the seven languages shared with Swivuriso.

\textbf{FLEURS:}~\cite{conneau-etal-2020-fleurs} is a multilingual speech corpus that covers 102 languages, created by recording text translated by humans into the respective languages. It includes only four South African languages, of which only isiXhosa and isiZulu were used in this work. We used test splits for both languages.

\subsection{Dataset Preprocessing}
All audio files were downsampled to 16 kHz. The transcriptions were preprocessed by normalizing them using Unicode Normalization Form C (NFC), removing all punctuation except a few. 
The retained punctuation includes ?!\'-,.;\%=+*\# and numeric digits.
We also removed special tokens we noticed in the data set, including \texttt{[pause]}, \texttt{[um]}, \texttt{[cs]}, and \texttt{[?]}, and lowercase all transcriptions.

\subsection{Model Architectures}
For our experiments, we use encoder-only ASR models based on Mamba or Conformer architectures, both trained with Connectionist Temporal Classification (CTC)~\cite{graves2006connectionist}, and use character-level modeling. We chose CTC due to its simplicity, computational efficiency, and previous use in related work~\cite{Flynn2026BeyondTU}. To ensure a fair comparison, both architectures share the same configuration, including 18 encoder layers, a hidden size of 512, and a feed-forward dimension of 2048, resulting in comparable model scales (\emph{114M} parameters for Conformer and \emph{123M} parameters for Mamba).

The Conformer follows the standard architecture~\cite{gulati20_interspeech}, consisting of relative positional self-attention~\cite{dai-etal-2019-transformer}, convolution modules, and feed-forward networks. In contrast, Mamba entirely replaces self-attention with a state-space sequence modeling mechanism. For the Mamba-based system, we adopt ConMamba~\cite{jiang2024speechslytherin}, an architecture designed for speech processing that integrates a bidirectional Mamba module (BiMamba; $d_{\text{state}}=16$, expand=2, $d_{\text{conv}}=4$), a feed-forward network and a convolutional module to efficiently capture local and long-range dependencies. 

\subsection{ASR Training Strategies}
\label{sec:train_strat}
Our experiments consider both \textbf{monolingual} and \textbf{multilingual} training regimes. In the monolingual setting, we train a separate model for each language using language-specific data and evaluate it in the corresponding language. For the multilingual setting, we evaluate five training strategies by combining monolingual data.


\begin{enumerate}
    \item \textbf{Multilingual-Implicit (MI)}: A single model trained on the pooled data from all languages without any explicit language information.

    \item \textbf{Multilingual-Implicit Family (MIF)}: For each language family, one model trained on pooled data from all languages in that family, without using any explicit language-identifying information.
    
    \item \textbf{Multilingual Language Embedding (MLE)}: Similar to (1), but with the introduction of a learnable language embedding matrix $E^{(\ell)} \in \mathbb{R}^{n_{\text{lang}} \times 32}$. For each input sequence that belongs to language $\ell$, the corresponding embedding is added to the acoustic representations produced by the CNN downsampling module:
    \begin{equation}
        \mathbf{h}'_t = \mathbf{h}_t + \mathbf{e}_{\ell}, \quad \forall t,
    \end{equation}
    where $\mathbf{h}_t \in \mathbb{R}^{32}$ is the frame-level acoustic representation and
    $\mathbf{e}_{\ell} \in \mathbb{R}^{32}$ is the embedding corresponding to language $\ell$.
    The resulting representations are then fed into the Mamba module. Unlike \cite{Tian2022LAELE}, we do not concatenate language embeddings to acoustic features.
    

    \item \textbf{Multilingual Language-Family Embedding (MLFE)}: We extend MLE by additionally introducing a language-family embedding matrix $E^{(f)} \in \mathbb{R}^{n_{\text{fam}} \times 32}$. The acoustic representations are augmented as:
    \begin{equation}
        \mathbf{h}'_t = \mathbf{h}_t + \mathbf{e}_{\ell} + \mathbf{e}_{f}, \quad \forall t,
    \end{equation}
    where $\mathbf{e}_{f}$ corresponds to the embedding of the language family.

    \item \textbf{Multilingual ASR + LID (M-CTC+LID)}: A multitask learning setup where a shared encoder is trained jointly with a CTC-based ASR objective and an auxiliary language identification (LID) head. The total loss is:
    \begin{equation}
        \mathcal{L} = \mathcal{L}_{\text{CTC}} + \lambda \mathcal{L}_{\text{LID}},
    \end{equation}
    where $\lambda$ controls the contribution of the LID objective.\footnote{$\lambda$ was set at 0.1 after ablation over {0.05, 0.1, 0.2, 0.3}.}
\end{enumerate}

\subsection{Training Configuration and Hyperparameters}
All models were trained with three seeds for 250 (monolingual) or 100 (multilingual) epochs on NVIDIA A100 GPUs (40GB or 80GB, with 80GB used for Conformer models). Training used \texttt{bfloat16} mixed-precision, a batch size of 32, and identical optimization settings following a publicly available SpeechBrain-based recipe~\cite{speechbrain}.\footnote{\url{https://github.com/mattmireles/Mamba-ASR}} We used the AdamW optimizer tuning the learning rate over {1e-3, 8e-4, 5e-4, 2e-4, 1e-4}.
All models were trained using a vocabulary of 189 characters derived from the Swivuriso combined training split. Although the target South African languages were not expected to require such a large character inventory, we observed that the dataset includes multiple characters with diacritics, as well as additional non-standard symbols, likely due to portions of the data being sourced from the web (Wikipedia). These include, among others, characters such as ß and Greek letters (e.g. $\sigma$ and $\lambda$).
For evaluation, we averaged the weights of the five best checkpoints per seed and report the mean performance over three seeds, with punctuation removed and Word Error Rate (WER) as the metric.

\begin{table*}[ht]
    \centering
        \caption{Performance comparison (WER \%) between monolingual Conformer and ConMamba models on long and short utterances. Each model is evaluated on the same language on which it was trained on (Swivuriso). Scores are averaged over three seeds, with standard deviations reported as subscripts.}
        \vspace{-8pt} 
    \begin{tabular}{lcccccccccc}
    \midrule
\rowcolor{white}
\cellcolor{white} & \cellcolor{white} 
& \multicolumn{3}{c}{\cellcolor{OIblue!30}\textbf{Nguni}} 
& \cellcolor{OIblue!80!OIorange!20}\textbf{Tsonga} 
& \multicolumn{2}{c}{\cellcolor{OIgreen!35}\textbf{Sotho}} 
& \cellcolor{OIgreen!80!OIvermillion!20}\textbf{Venda} \\
   
    \textbf{Setup} &  \textbf{Size}	& \textbf{nbl} & \textbf{xho} &  \textbf{zul} & \textbf{tso}	& \textbf{sot} &	\textbf{tsn}	 &	\textbf{ven} 		& \textbf{Avg} \\
    \midrule
    \multicolumn{9}{l}{\textbf{\textit{Short Utterances (0-30s)}}}\\
    Conformer & 114M &  42.29$_{2.08}$ & 40.28$_{0.55}$ & 45.08$_{1.11}$ & 35.19$_{3.68}$ & 31.09$_{1.64}$ & 26.76$_{2.96}$ & 27.71$_{1.59}$ & 35.49 \\
    ConMamba & 123M & 40.72$_{1.19}$ & 40.18$_{1.27}$ & 44.19$_{0.99}$ & 29.92$_{1.04}$ & 29.25$_{0.10}$ & 23.30$_{0.06}$ & 22.77$_{0.14}$ & 32.91 \\
    \midrule
    \multicolumn{9}{l}{\textbf{\textit{Long Utterances ($>$30s)}}}\\
    Conformer & 114M & 43.18$_{2.51}$ & 38.26$_{0.54}$ & 47.36$_{1.16}$ & 41.40$_{3.62}$ & 31.54$_{1.55}$ & 32.12$_{3.66}$ & 31.69$_{1.73}$ & 37.94 \\
    ConMamba & 123M & 42.11$_{1.14}$ & 38.66$_{1.46}$ & 47.09$_{1.14}$ & 35.94$_{1.00}$ & 29.70$_{0.12}$ & 28.25$_{0.12}$ & 25.98$_{0.16}$ & 35.39 \\
    
    \bottomrule
    \end{tabular}
    \label{tab:monolingual}
    \vspace{-8pt} 
\end{table*}

\section{Result and Discussion}
In this section, we present the performance of Mamba for ASR across seven languages. We first compare Mamba with Conformer in a monolingual setting, followed by an evaluation of both models on long-context ASR. We then assess Mamba's effectiveness in multilingual training by comparing five training strategies, examine its cross-dataset generalization, and finally present an ablation study of the multilingual Mamba model with language embeddings (MLE).

\subsection{Monolingual ASR Performance}
\Cref{tab:monolingual} presents the monolingual WER comparison between Conformer and Mamba models on both short and long utterances. All models were trained on short utterances and evaluated on the same language.
The in-domain evaluation result shows comparable performance across languages for Conformer and ConMamba: on average, ConMamba achieves a 33.43\% WER, while Conformer achieves 35.49\%. Despite comparable accuracy, ConMamba, with nearly 10M additional parameters, is more efficient in training, inference, and memory usage~\cite{zevallos25_interspeech}. In our experiments, Mamba required approximately 18 hours of training per language, compared to 34 hours per language for Conformer. Due to higher memory requirements, Conformer could not be trained on A100 40GB GPUs and therefore required A100 80GB GPUs, whereas Mamba could be trained on A100 40GB GPUs. Furthermore, the results indicate that the Nguni languages (\texttt{nbl}, \texttt{xho}, and \texttt{zul}) consistently yield the lowest performance across both models, followed by \texttt{tso}, a related language. In contrast, \texttt{ven} and \texttt{tsn} achieve the strongest results across both models. All monolingual models were trained under identical experimental conditions, making data imbalance an unlikely explanation. We hypothesize that this pattern may be linked to the complex agglutinative morphology of Nguni languages. Furthermore, not including English in our training could affect performance, especially in code-switching scenarios where English is often used~\cite{van-der-westhuizen-niesler-2018-first}. 
More analysis is needed to better understand these challenges and improve performance.

\subsection{Robustness to Length Variation}

Evaluation on long utterances outside the training length distribution shows that WER is similar to or slightly higher than for short utterances in most languages, except \texttt{xho}, where it is lower. The average WER difference of $\sim2.5\%$ indicates comparable length generalization across models, with similar degradation trends likely due to the CTC decoder. In ~\cite{miyazaki24_interspeech} it was shown that incorporating Mamba in encoder-decoder architectures helps reduce degradation for long-form ASR, and we investigate if the same holds true for encoder-only ASR.

To investigate this, we performed an ablation experiment by sampling and concatenating \emph{dev\_test} utterances of up to 60s, inserting 0.12 s of silence between them, to create sequences spanning progressively longer ranges (i.e. 60-90s, 90-120s, \dots, 210-240s). Scripted and unscripted speech from the same speaker was kept separate and concatenation was restricted to utterances from that speaker.

 \begin{figure}[t]
    \centering
    \includegraphics[width=0.45\textwidth]{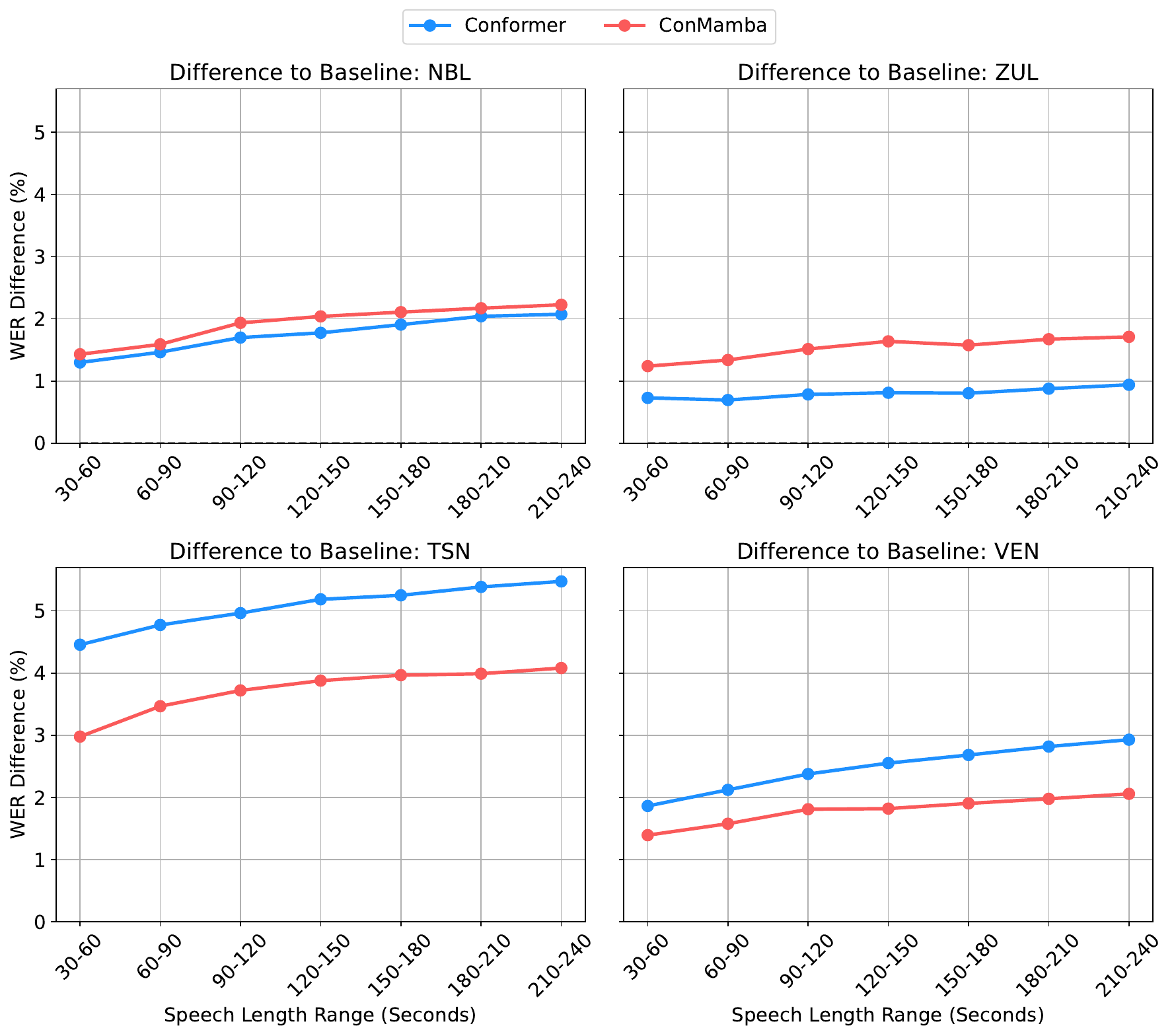}
    \caption{Difference in WER (\%) relative to baseline for Conformer and ConMamba across increasing speech length ranges.}
    \label{fig:length_wer_diffs}
    \vspace{-10pt} 
\end{figure}

\Cref{fig:length_wer_diffs} shows the performance degradation of the monolingual models for \texttt{nbl}, \texttt{tsn}, \texttt{ven}, and \texttt{zul}, using inference on the original short utterances (0–30 s) as baselines. We observe that both architectures experience increased degradation as the utterance duration grows; however, ConMamba exhibits slightly less degradation overall compared to Conformer. Neither model is fully robust to longer utterances, and the degradation patterns appear similar for both architectures. Future work should evaluate these architectures on authentic long utterances to better assess their robustness as we already observe degradation when simply concatenating short utterances into ranges of at least 30s.

Thus far, we have shown that Mamba matches Conformer across all seven languages while being more computationally efficient, although both models degrade on long-form ASR. We evaluate Mamba in the multilingual setting by comparing training strategies and assessing cross-corpus generalization.

\begin{table*}[ht]
    \centering
    \caption{Performance (WER \%) of multilingual Mamba models trained under different regimes on individual languages (top), evaluated on the full Swivuriso dev-test set. The bottom section reports generalization to NCHLT. Results include the originally trained monolingual baselines. Scores are averaged over three seeds; standard deviations are shown as subscripts.} \vspace{-8pt} 
    \begin{tabular}{lccccccccc}
    \midrule
    
   \rowcolor{white} &  \multicolumn{3}{c}{\cellcolor{OIblue!30}\textbf{Nguni}} 
& \cellcolor{OIblue!80!OIorange!20}\textbf{Tsonga} 
& \multicolumn{2}{c}{\cellcolor{OIgreen!35}\textbf{Sotho}} 
& \cellcolor{OIgreen!80!OIvermillion!20}\textbf{Venda} \\
   
    \textbf{Setup}	& \textbf{nbl} & \textbf{xho} &  \textbf{zul} & \textbf{tso}	& \textbf{sot} &	\textbf{tsn}	 &	\textbf{ven} 		& \textbf{Avg} \\
    \midrule
    \multicolumn{9}{c}{\textbf{Swivuriso $\rightarrow$ Swivuriso}}\\
    \rowcolor{LightAsh} \multicolumn{9}{l}{\textbf{\textit{Monolingual}}}\\
    Conformer & 42.55$_{2.20}$ & 39.16$_{0.53}$ & 46.54$_{1.10}$ & 37.96$_{3.68}$ & 31.32$_{1.59}$ & 29.26$_{3.30}$ & 29.69$_{1.64}$ & 36.64 \\
ConMamba & 41.18$_{1.15}$ & 39.34$_{1.36}$ & 46.05$_{1.09}$ & 32.59$_{1.01}$ & 29.48$_{0.11}$ & 25.61$_{0.07}$ & 24.42$_{0.14}$ & 34.10 \\

    \rowcolor{OIblue!15}\multicolumn{9}{l}{\textbf{\textit{Multilingual (ConMamba)}}}\\
    MIF & 36.41$_{0.85}$ & 35.80$_{1.12}$ & 41.94$_{1.10}$ & 30.58$_{1.50}$  & 28.22$_{1.82}$ & 23.37$_{1.79}$ & 23.72$_{1.58}$ & 31.43 \\
    MI & 34.76$_{0.39}$ & 34.05$_{0.20}$ & 40.33$_{0.31}$ & 27.47$_{0.44}$ & 26.41$_{0.21}$ & 21.58$_{0.28}$ & 21.69$_{0.14}$ & 29.47 \\
    MLE & \textbf{34.27}$_{0.07}$ & \textbf{33.71}$_{0.13}$ & \textbf{39.85}$_{0.11}$ & \textbf{26.85}$_{0.08}$ & \textbf{26.11}$_{0.06}$ & \textbf{21.28}$_{0.11}$ & \textbf{21.35}$_{0.08}$ & \textbf{29.06} \\
    MLFE & 34.37$_{0.24}$ & 33.88$_{0.12}$ & 40.15$_{0.15}$ & 27.09$_{0.32}$ & 26.22$_{0.11}$ & 21.46$_{0.19}$ & 21.68$_{0.17}$ & 29.26 \\
    M-CTC+LID & 36.28$_{0.47}$ & 35.62$_{0.34}$ & 41.77$_{0.42}$ & 29.52$_{0.56}$ & 27.45$_{0.59}$ & 22.73$_{0.33}$ & 22.73$_{0.17}$ & 30.87 \\
    \midrule
    \multicolumn{9}{c}{\textbf{Swivuriso $\rightarrow$ NCHLT}}\\
    \rowcolor{LightAsh} \multicolumn{9}{l}{\textbf{\textit{Monolingual}}}\\
    Conformer & 47.75$_{3.45}$ & 46.75$_{1.47}$ & 48.20$_{1.28}$ & 28.00$_{2.24}$ & 43.06$_{5.21}$ & 40.20$_{6.67}$ & 51.25$_{2.16}$ & 43.60 \\
    ConMamba & 44.09$_{0.80}$ & 47.56$_{3.07}$ & 48.30$_{1.50}$ & 21.34$_{0.64}$ & 35.67$_{0.61}$ & 26.46$_{0.56}$ & 38.79$_{0.73}$ & 37.46 \\
    \rowcolor{OIblue!15}\multicolumn{9}{l}{\textbf{\textit{Multilingual (ConMamba)}}}\\
    MIF & 35.33$_{0.80}$ & 42.37$_{2.13}$ & 42.78$_{2.17}$ & 21.44$_{1.56}$ & 35.37$_{3.63}$ & 23.66$_{2.13}$ & 37.40$_{3.78}$  & 34.05 \\
    MI & 33.22$_{0.67}$ & 39.16$_{1.22}$ & 40.09$_{1.16}$ & 19.77$_{0.50}$ & 33.44$_{1.54}$ & 21.07$_{1.00}$ & 33.68$_{1.57}$ & 31.49 \\
    MLE & \textbf{30.71}$_{0.59}$ & \textbf{36.73}$_{1.15}$ & \textbf{36.42}$_{0.38}$ & \textbf{17.44}$_{0.31}$ & \textbf{30.99}$_{2.06}$ & \textbf{19.43}$_{1.20}$ & \textbf{29.58}$_{0.59}$ & \textbf{28.76} \\
    MLFE & 30.80$_{0.11}$ & 36.77$_{0.33}$ & 36.33$_{1.64}$ & 17.25$_{0.07}$ & 31.49$_{1.87}$ & 20.63$_{1.64}$ & 30.01$_{0.44}$ & 29.04 \\
    M-CTC+LID & 35.01$_{1.16}$ & 40.85$_{0.82}$ & 40.78$_{3.05}$ & 20.72$_{1.06}$ & 36.01$_{2.76}$ & 22.59$_{1.85}$ & 37.02$_{0.35}$ & 33.28 \\
    
    \bottomrule
    \end{tabular}
    \label{tab:multilingual}
    \vspace{-8pt} 
\end{table*}
\begin{table}[h]
    \centering
    \caption{Generalization performance (WER \%) of multilingual Mamba models trained under different training regimes on FLEURS, compared against the monolingual baseline models.} \vspace{-8pt} 
    \begin{tabular}{lccc}
    \midrule
    \textbf{Setup}	& \cellcolor{OIblue!30}\textbf{xho}	& \cellcolor{OIblue!30}\textbf{zul}	& \textbf{Avg} \\
    \midrule
    \multicolumn{4}{c}{\textbf{Swivuriso $\rightarrow$ FLEURS}}\\
    \rowcolor{LightAsh}\multicolumn{4}{l}{\textbf{\textit{Monolingual}}}\\
    Conformer & 55.57$_{0.58}$ & 46.97$_{1.16}$ & 51.27 \\
    ConMamba & 56.26$_{2.50}$ & 45.64$_{1.46}$ & 50.95 \\
    \rowcolor{OIblue!15}\multicolumn{4}{l}{\textbf{\textit{Multilingual (ConMamba)}}}\\
    MIF & 51.14$_{1.99}$ & 39.88$_{1.67}$ & 45.51 \\
    MI & 48.29$_{0.33}$ & 37.49$_{0.78}$ & 42.89 \\
    MLE & \textbf{47.03}$_{0.10}$ & \textbf{36.59}$_{0.26}$ & \textbf{41.81} \\
    MLFE  & 47.95$_0.43$ & 37.16$_0.19$ & 42.56 \\
    M-CTC+LID & 49.98$_{0.35}$ & 39.35$_{0.28}$ & 44.66 \\  
    \bottomrule
    \end{tabular}
    \label{tab:tofleurs}
    \vspace{-8pt} 
\end{table}

\subsection{Multilingual Mamba ASR Performance}
\Cref{tab:multilingual} presents results for multilingual models trained on short utterances from the seven Swivuriso languages and evaluated on the Swivuriso \emph{dev\_test} sets. We consider five multilingual training regimes described in \Cref{sec:train_strat}, including MIF, which serves as a multilingual baseline. In MIF, Mamba models are trained on the Nguni-Tsonga and Sotho-Venda language groupings, respectively. All multilingual models outperform their monolingual counterparts. The smallest improvement is achieved by MIF, with an average WER reduction of 2.68\% relative to the monolingual baseline, while the fully multilingual multitask model achieves a reduction of 4\%. Explicit language conditioning (MLE and MLFE) yields only marginal improvements over multilingual implicit training (MI) in terms of final WER. MLE and MLFE converge faster, achieving up to 7\% lower validation error than MI during the early training epochs. MI reaches comparable validation performance after approximately 20 epochs.

At the language level, comparing MLE with the monolingual model, the Nguni languages benefit more, with \texttt{nbl} improving by $\sim7\%$, \texttt{xho} by $\sim5\%$, and \texttt{zul} by $\sim6\%$, while the Sotho languages show smaller gains of $\sim3\%$ and $\sim4\%$, possibly due to their stronger monolingual performance. The higher gains for Nguni languages may be attributed to the additional training data available from related languages.

\subsection{Cross-Corpus Robustness Performance}
Comparing multilingual and monolingual models on cross-corpus performance, \Cref{tab:multilingual,tab:tofleurs} shows the results on NCHLT and FLEURS, respectively. On both datasets, we observe gaps of 8.7\% and 9.14\% between the monolingual ConMamba model and the best-performing multilingual model (MLE). The language-conditioned model also outperforms the other multilingual models, with improvements of 2.73\% and 1.08\% over the MI model on NCHLT and FLEURS, respectively. Overall, multilingual models demonstrate better cross-corpus robustness.

Thus far, we have observed that while language conditioning does not yield large in-domain improvements, it leads to stronger gains in cross-corpus generalization. Hence, we investigate this effect through an ablation study of the multilingual model with language embeddings.

\begin{table*}[ht]
    \centering
    \caption{Performance (WER, \%) of multilingual Mamba models trained on 50-hour per-language data after language embedding ablations (zeroing and permutation). Evaluated on the Swivuriso dev-test set. Results are averaged over three seeds; subscripts denote standard deviations.} \vspace{-8pt} 
    \begin{tabular}{lccccccccc}
    \midrule
    
   \rowcolor{white} &  \multicolumn{3}{c}{\cellcolor{OIblue!30}\textbf{Nguni}} 
& \cellcolor{OIblue!80!OIorange!20}\textbf{Tsonga} 
& \multicolumn{2}{c}{\cellcolor{OIgreen!35}\textbf{Sotho}} 
& \cellcolor{OIgreen!80!OIvermillion!20}\textbf{Venda} \\
   
    \textbf{Setup}	& \textbf{nbl} & \textbf{xho} &  \textbf{zul} & \textbf{tso}	& \textbf{sot} &	\textbf{tsn}	 &	\textbf{ven} 		& \textbf{Avg} \\
    \midrule
    MLE &  34.27$_{0.07}$ & 33.71$_{0.13}$ & 39.85$_{0.11}$ & 26.85$_{0.08}$ & 26.11$_{0.06}$ & 21.28$_{0.11}$ & 21.35$_{0.08}$ & 29.06 \\
    Zeroed &  67.42$_{27.72}$ & 57.83$_{22.85}$ & 57.97$_{13.38}$ & 74.34$_{25.08}$ & 60.10$_{30.50}$ & 54.44$_{35.76}$ & 69.21$_{32.97}$ & 63.04 \\
    Permuted &  129.18$_{59.70}$ & 63.70$_{37.44}$ & 117.93$_{51.87}$ & 82.33$_{9.07}$ & 56.86$_{29.80}$ & 46.89$_{30.16}$ & 61.11$_{33.16}$ & 79.72  \\

    \bottomrule
    \end{tabular}
    \label{tab:emb-ablate}
    \vspace{-8pt} 
\end{table*}
\begin{table*}[ht]
    \centering
    \caption{Performance (WER, \%) of multilingual Mamba models under 5-hour and 10-hour per-language training. Evaluated on the Swivuriso dev-test set. Results are averaged over three seeds; subscripts denote standard deviations.} \vspace{-8pt} 
    \begin{tabular}{lccccccccc}
    \midrule
    
   \rowcolor{white}  & \multicolumn{3}{c}{\cellcolor{OIblue!30}\textbf{Nguni}} 
& \cellcolor{OIblue!80!OIorange!20}\textbf{Tsonga} 
& \multicolumn{2}{c}{\cellcolor{OIgreen!35}\textbf{Sotho}} 
& \cellcolor{OIgreen!80!OIvermillion!20}\textbf{Venda} \\
   
    \textbf{Setup}	& \textbf{nbl} & \textbf{xho} &  \textbf{zul} & \textbf{tso}	& \textbf{sot} &	\textbf{tsn}	 &	\textbf{ven} 		& \textbf{Avg} \\
    \midrule
    \rowcolor{LightAsh} \multicolumn{9}{l}{\textbf{\textit{5 hours} (\textbf{Swivuriso $\rightarrow$ Swivuriso})}}\\
    MI & 63.71$_{0.31}$ & 63.25$_{0.30}$ & 66.63$_{0.20}$ & 62.45$_{0.59}$ & 51.94$_{0.19}$ & 48.15$_{0.38}$ & 47.13$_{0.57}$ & 57.61 \\
    MLE &  \textbf{61.08}$_{0.24}$ & \textbf{61.48}$_{0.19}$ & \textbf{65.46}$_{0.34}$ & \textbf{59.80}$_{0.30}$ & \textbf{50.18}$_{0.37}$ & \textbf{46.32}$_{0.28}$ & \textbf{45.80}$_{0.64}$ & \textbf{55.73} \\
    \midrule
$\Delta_{\mathrm{MI-MLE}}$ & 2.63 & 1.77 & 1.17 & 2.65 & 1.75 & 1.83 & 1.32 & 1.88 \\
    \rowcolor{LightAsh} \multicolumn{9}{l}{\textbf{\textit{10 hours} (\textbf{Swivuriso $\rightarrow$ Swivuriso})}}\\
    MI & 49.67$_{0.21}$ & 49.95$_{0.32}$ & 54.58$_{0.26}$ & 46.67$_{0.29}$ & 38.81$_{0.35}$ & 34.21$_{0.26}$ & 33.67$_{0.39}$ & 43.94 \\
    MLE & \textbf{48.03}$_{0.20}$ & \textbf{48.15}$_{0.07}$ & \textbf{53.16}$_{0.04}$ & \textbf{44.53}$_{0.18}$ & \textbf{37.49}$_{0.23}$ & \textbf{33.05}$_{0.23}$ & \textbf{32.57}$_{0.15}$ & \textbf{42.42} \\
    \midrule
$\Delta_{\mathrm{MI-MLE}}$ & 1.64 & 1.80 & 1.41 & 2.15 & 1.32 & 1.15 & 1.11 & 1.51 \\

    \bottomrule
    \end{tabular}
    \label{tab:low-resource}
    \vspace{-8pt} 
\end{table*}

\subsection{Effect of Language Embedding Ablation}
To assess whether the learned language embeddings actively contribute to model predictions, we perform an ablation study where the language conditioning embedding is removed at inference time by replacing it with either a zero vector or a permuted version of the original embedding (to preserve magnitude while disrupting its structure). This allows us to isolate the contribution of the language embeddings from the rest of the model, even though performance improvements on in-domain data were not observed compared to training without any language conditioning. We present the results in~\Cref{tab:emb-ablate}. The results show that both ablations degrade performance across all languages, with permutation causing a greater degradation. This confirms the that the embeddings learned for the languages are informative.

\subsection{Effect of Language Conditioning in Low-Resource Regimes}
Given the comparable performance between MI and MLE in multilingual in-domain settings, as well as the observed robustness of MLE on out-of-domain datasets such as FLUERS and NCHLT, we hypothesize that language conditioning may improve performance in low-data scenarios. To verify this, we train MI and MLE models on 5-hour and 10-hour per-language subsets sampled from Swivuriso. \Cref{tab:low-resource} shows the results of this synthetic low-resource regime. MLE consistently outperforms MI in most languages in both data settings, yielding at least 1\% absolute WER reduction, with the average improvement being slightly higher in the 5-hour setting. This shows that language conditioning is particularly beneficial under low-resource conditions.

\begin{figure}[t]
    \centering    
    \includegraphics[width=0.9\linewidth]{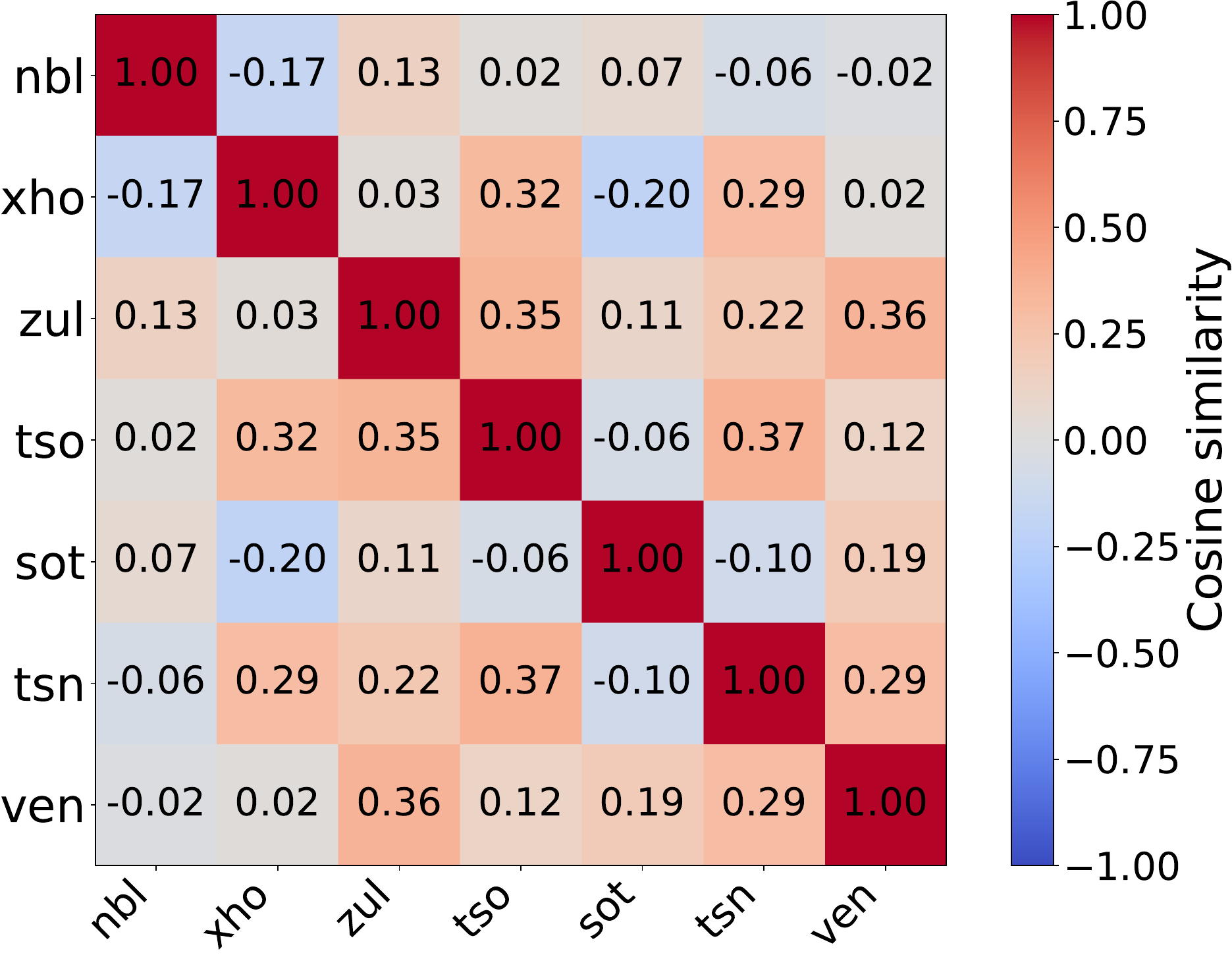}
    \caption{Cosine Similarity of the embeddings from 50-hour per language MLE.}  
    \label{fig:language-matrix}
    \vspace{-10pt}
\end{figure}

\subsection{Analysis of Learned Language Embeddings}
Given the importance of the learned language embeddings, a natural question is whether they encode linguistic information shared across languages. To investigate this, we extract the embedding matrix (for the 50-hour MLE model) and compute pairwise cosine similarities to analyze the structure of the resulting space. \Cref{fig:language-matrix} shows the resulting cosine similarity matrix, where we observe that the embedding space does not strictly reflect linguistic relatedness: linguistically similar languages do not consistently exhibit higher similarity, while some unrelated languages appear closer. This suggests that the embeddings do not encode typological similarity, but instead function as task-specific control vectors that steer the shared encoder representations to optimize the CTC objective.

\section{Conclusion}
In this work, we systematically evaluate the Mamba architecture for ASR in seven African languages. We first compare Mamba with Conformer in monolingual settings and assess robustness to long-context speech. We then explore multilingual training strategies, including pooled data, language embeddings, hierarchical language-family embeddings, and multitask learning with an auxiliary LID objective. Our results show that Mamba achieves competitive performance while being more computationally efficient than Conformer. Multilingual training consistently outperforms monolingual models, while explicit language conditioning does not yield consistent gains in the high-resource setting yet benefits low-resource and cross-corpus settings. We further find that learned language embeddings do not reflect typological similarity but instead act as task-specific control vectors. To the best of our knowledge, no publicly available multilingual pretrained Mamba speech encoders support the studied languages, limiting direct comparison to prior work. Future work should explore larger-scale pretraining and more expressive conditioning strategies for multilingual and low-resource speech models.

\section{Acknowledgments}JOA and BMA are funded by the Deutsche Forschungsgemeinschaft (DFG, German Research Foundation) – Project-ID 232722074 – SFB 1102.

\section{Generative AI Use Disclosure}
Generative AI tools were used exclusively for editing and polishing assistance, coding support for training and evaluation modules, code debugging, and manuscript proofreading including grammar and typographical corrections. 
No part of the scientific content, experimental design, analysis, citations, or conclusions was generated by AI tools. 

\bibliographystyle{IEEEtran}
\bibliography{mybib}

\begin{thebibliography}{10}
\providecommand{\url}[1]{#1}
\csname url@samestyle\endcsname
\providecommand{\newblock}{\relax}
\providecommand{\bibinfo}[2]{#2}
\providecommand{\BIBentrySTDinterwordspacing}{\spaceskip=0pt\relax}
\providecommand{\BIBentryALTinterwordstretchfactor}{4}
\providecommand{\BIBentryALTinterwordspacing}{\spaceskip=\fontdimen2\font plus
\BIBentryALTinterwordstretchfactor\fontdimen3\font minus \fontdimen4\font\relax}
\providecommand{\BIBforeignlanguage}[2]{{%
\expandafter\ifx\csname l@#1\endcsname\relax
\typeout{** WARNING: IEEEtran.bst: No hyphenation pattern has been}%
\typeout{** loaded for the language `#1'. Using the pattern for}%
\typeout{** the default language instead.}%
\else
\language=\csname l@#1\endcsname
\fi
#2}}
\providecommand{\BIBdecl}{\relax}
\BIBdecl

\bibitem{radford2023robust}
A.~Radford, J.~W. Kim, T.~Xu, G.~Brockman, C.~McLeavey, and I.~Sutskever, ``Robust speech recognition via large-scale weak supervision,'' in \emph{International conference on machine learning}, {Baltimore, USA}, 2023.

\bibitem{kim2023branchformer}
K.~Kim, F.~Wu, Y.~Peng, J.~Pan, P.~Sridhar, K.~J. Han, and S.~Watanabe, ``E-branchformer: Branchformer with enhanced merging for speech recognition,'' in \emph{IEEE Spoken Language Technology Workshop (SLT)}, 2022.

\bibitem{prabhavalkar2023end}
R.~Prabhavalkar, T.~Hori, T.~N. Sainath, R.~Schl{\"u}ter, and S.~Watanabe, ``End-to-end speech recognition: A survey,'' \emph{IEEE/ACM Transactions on Audio, Speech, and Language Processing}, vol.~32, pp. 325--351, 2023.

\bibitem{gulati20_interspeech}
A.~Gulati, J.~Qin, C.-C. Chiu, N.~Parmar, Y.~Zhang, J.~Yu, W.~Han, S.~Wang, Z.~Zhang, Y.~Wu, and R.~Pang, ``{Conformer: Convolution-augmented Transformer for Speech Recognition},'' in \emph{{Interspeech}}, Virtual, 2020.

\bibitem{gu2024mamba}
A.~Gu and T.~Dao, ``Mamba: Linear-time sequence modeling with selective state spaces,'' in \emph{First Conference on Language Modeling}, Philadelphia, USA, 2024.

\bibitem{zevallos25_interspeech}
R.~Zevallos, M.~{Cortada Garcia}, S.~Solito, C.~Mena, A.~Peiró-Lilja, and J.~Hernando, ``{Assessing the Performance and Efficiency of Mamba ASR in Low-Resource Scenarios},'' in \emph{{Interspeech}}, Rotterdam, The Netherlands, 2025.

\bibitem{moriya25_interspeech}
T.~Moriya, M.~Mimura, K.~Matsui, H.~Sato, and K.~Matsuura, ``{Attention-Free Dual-Mode ASR with Latency-Controlled Selective State Spaces},'' in \emph{{Interspeech}}, Rotterdam, The Netherlands, 2025.

\bibitem{Ali2025MLMATM}
M.~N. Ali, D.~Falavigna, and A.~Brutti, ``Mlma: Towards multilingual asr with mamba-based architectures,'' \emph{ArXiv}, vol. abs/2510.18684, 2025.

\bibitem{ahia-etal-2021-low-resource}
O.~Ahia, J.~Kreutzer, and S.~Hooker, ``The low-resource double bind: An empirical study of pruning for low-resource machine translation,'' in \emph{Findings of the Association for Computational Linguistics: Empirical Methods in Natural Language Processing (EMNLP)}, Punta Cana, Dominican Republic, 2021.

\bibitem{marivate2025swivuriso0}
V.~Marivate, K.~Olaleye, S.~Mundia, A.~Bakainga, U.~Netshifhefhe, M.~Milanzie, T.~H. Mogale, T.~Sindane, Z.~Abdulrasaq, K.~Mokgosi, C.~Okorie, N.~Z.~V. Wyk, G.~Morrissey, D.~Dunbar, F.~Smit, T.~Chidi, R.~Mabuya, A.~Bukula, R.~Mlambo, T.~Macucwa, I.~Abdulmumin, , and S.~Rananga, ``Swivuriso: The south african next voices multilingual speech dataset,'' \emph{arXiv preprint arXiv: 2512.02201}, 2025.

\bibitem{Diack2026WAXALAL}
A.~D. Diack, P.~H. Nelson, K.~Agbesi, A.~Nakalembe, M.~Mohamedkhair, V.~Dube, T.~Siyavora, S.~Venugopalan, J.~Hickey, U.~Okonkwo, A.~Bapna, I.~Wiafe, R.~D. Helegah, E.~D. Atsakpo, C.~Nutrokpor, F.~B.~P. Winful, K.~K. Solaga, J.-D. Abdulai, A.~O. Ekpezu, A.~Niyonkuru, S.~Rutunda, B.~Ishimwe, M.~Melese, E.~Bainomugisha, J.~Nakatumba‐Nabende, A.~Katumba, C.~Babirye, J.~Mukiibi, V.~Kimani, S.~Kibacia, J.~Maina, F.~Emmah, A.~I. Shekarau, I.~Adamu, Y.~S. Abdullahi, H.~Lakougna, B.~Macdonald, H.~Shemtov, A.~Walcott-Bryant, M.~Ciss{\'e}, A.~Hassidim, J.~Dean, and Y.~Matias, ``Waxal: A large-scale multilingual african language speech corpus,'' 2026.

\bibitem{Flynn2026BeyondTU}
R.~Flynn and A.~Ragni, ``Beyond the utterance: An empirical study of very long context speech recognition,'' \emph{IEEE Transactions on Audio, Speech and Language Processing}, vol.~34, pp. 910--920, 2026.

\bibitem{alabi-etal-2025-charting}
J.~O. Alabi, M.~A. Hedderich, D.~I. Adelani, and D.~Klakow, ``Charting the landscape of {A}frican {NLP}: Mapping progress and shaping the road ahead,'' in \emph{Conference on Empirical Methods in Natural Language Processing (EMNLP)}, Suzhou, China, 2025.

\bibitem{imam-etal-2025-automatic}
S.~H. Imam, B.~Sani, D.~K. Gete, B.~Y. Ahmed, I.~S. Ahmad, I.~Abdulmumin, S.~M. Yimam, M.~Y. Bello, and S.~H. Muhammad, ``Automatic speech recognition for {A}frican low-resource languages: Challenges and future directions,'' in \emph{6th Workshop on African Natural Language Processing (AfricaNLP)}, Vienna, Austria, 2025.

\bibitem{van-der-westhuizen-niesler-2018-first}
E.~van~der Westhuizen and T.~Niesler, ``A first {S}outh {A}frican corpus of multilingual code-switched soap opera speech,'' in \emph{11th International Conference on Language Resources and Evaluation ({LREC})}, Miyazaki, Japan, 2018.

\bibitem{barnard14_sltu}
E.~Barnard, M.~H. Davel, C.~van Heerden, F.~de~Wet, and J.~Badenhorst, ``{The NCHLT speech corpus of the South African languages},'' in \emph{{4th Workshop on Spoken Language Technologies for Under-Resourced Languages (SLTU)}}, St. Petersburg, Russia, 2014.

\bibitem{BADENHORST2022107860}
J.~Badenhorst and F.~{de Wet}, ``Nchlt auxiliary speech data for asr technology development in south africa,'' \emph{Data in Brief}, vol.~41, p. 107860, 2022.

\bibitem{gumede-plauche-2009-initial}
T.~Gumede and M.~Plauch{\'e}, ``Initial fieldwork for {LWAZI}: A telephone-based spoken dialog system for rural {S}outh {A}frica,'' in \emph{1st Workshop on Language Technologies for {A}frican Languages}, Athens, Greece, 2009.

\bibitem{ardila-etal-2020-common}
R.~Ardila, M.~Branson, K.~Davis, M.~Kohler, J.~Meyer, M.~Henretty, R.~Morais, L.~Saunders, F.~Tyers, and G.~Weber, ``\BIBforeignlanguage{eng}{Common voice: A massively-multilingual speech corpus},'' in \emph{\BIBforeignlanguage{eng}{12th Language Resources and Evaluation Conference}}, Marseille, France, 2020.

\bibitem{conneau-etal-2020-fleurs}
A.~Conneau, M.~Ma, S.~Khanuja, Y.~Zhang, V.~Axelrod, S.~Dalmia, J.~Riesa, C.~Rivera, and A.~Bapna, ``Fleurs: Few-shot learning evaluation of universal representations of speech,'' in \emph{IEEE Spoken Language Technology Workshop (SLT)}, Doha, Qatar, 2022.

\bibitem{rajab-etal-2025-esethu}
J.~Rajab, A.~Aremu, E.~A. Chimoto, D.~Dunbar, G.~Morrissey, F.~Thior, L.~Potgieter, J.~Ojo, A.~L. Tonja, W.~N. Nekoto, P.~Moiloa, J.~Abbott, V.~Marivate, and B.~Rosman, ``The esethu framework: Reimagining sustainable dataset governance and curation for low-resource languages,'' in \emph{63rd Annual Meeting of the Association for Computational Linguistics (Volume 1: Long Papers)}, Vienna, Austria, 2025.

\bibitem{ylmaz18c_interspeech}
E.~Yilmaz, A.~Biswas, E.~{van der Westhuizen}, F.~{de Wet}, and T.~Niesler, ``{Building a Unified Code-Switching ASR System for South African Languages},'' in \emph{{Interspeech}}, {Hyderabad, India}, 2018, pp. 1923--1927.

\bibitem{BISWAS2022101262}
A.~Biswas, E.~Yilmaz, E.~{van der Westhuizen}, F.~{de Wet}, and T.~Niesler, ``Code-switched automatic speech recognition in five south african languages,'' \emph{Computer Speech \& Language}, vol.~71, p. 101262, 2022.

\bibitem{biswas19b_interspeech}
A.~Biswas, E.~Yılmaz, F.~de~Wet, E.~van~der Westhuizen, and T.~Niesler, ``{Semi-Supervised Acoustic Model Training for Five-Lingual Code-Switched ASR},'' in \emph{{Interspeech}}, {Graz, Austria}, 2019.

\bibitem{biswas-etal-2020-semi-supervised}
A.~Biswas, E.~Yilmaz, F.~de~Wet, E.~van~der Westhuizen, and T.~Niesler, ``\BIBforeignlanguage{eng}{Semi-supervised development of {ASR} systems for multilingual code-switched speech in under-resourced languages},'' in \emph{\BIBforeignlanguage{eng}{12th Language Resources and Evaluation Conference {(LREC)}}}, Marseille, France, 2020.

\bibitem{boito2024mhubert}
M.~{Zanon Boito}, V.~Iyer, N.~Lagos, L.~Besacier, and I.~Calapodescu, ``mhubert-147: A compact multilingual hubert model,'' in \emph{Interspeech}, Kos, Greece, 2024.

\bibitem{alabi25_interspeech}
J.~O. Alabi, X.~Liu, D.~Klakow, and J.~Yamagishi, ``{AfriHuBERT: A self-supervised speech representation model for African languages},'' in \emph{{Interspeech}}, Rotterdam, The Netherlands, 2025.

\bibitem{graves2006connectionist}
A.~Graves, S.~Fern\'{a}ndez, F.~Gomez, and J.~Schmidhuber, ``Connectionist temporal classification: labelling unsegmented sequence data with recurrent neural networks,'' in \emph{23rd International Conference on Machine Learning {(ICML)}}, New York, USA, 2006.

\bibitem{dai-etal-2019-transformer}
Z.~Dai, Z.~Yang, Y.~Yang, J.~Carbonell, Q.~Le, and R.~Salakhutdinov, ``Transformer-{XL}: Attentive language models beyond a fixed-length context,'' in \emph{57th Annual Meeting of the Association for Computational Linguistics}, Florence, Italy, 2019.

\bibitem{jiang2024speechslytherin}
X.~Jiang, Y.~A. Li, A.~N. Florea, C.~Han, and N.~Mesgarani, ``Speech slytherin: Examining the performance and efficiency of mamba for speech separation, recognition, and synthesis,'' in \emph{IEEE International Conference on Acoustics, Speech and Signal Processing (ICASSP)}, Hyderabad, India, 2025.

\bibitem{Tian2022LAELE}
J.~Tian, J.~Yu, C.~Zhang, C.~Weng, Y.~Zou, and D.~Yu, ``Lae: Language-aware encoder for monolingual and multilingual asr,'' in \emph{Interspeech}, Incheon, Korea, 2022.

\bibitem{speechbrain}
M.~Ravanelli, T.~Parcollet, P.~Plantinga, A.~Rouhe, S.~Cornell, L.~Lugosch, C.~Subakan, N.~Dawalatabad, A.~Heba, J.~Zhong, J.-C. Chou, S.-L. Yeh, S.-W. Fu, C.-F. Liao, E.~Rastorgueva, F.~Grondin, W.~Aris, H.~Na, Y.~Gao, R.~D. Mori, and Y.~Bengio, ``{SpeechBrain}: A general-purpose speech toolkit,'' 2021, arXiv:2106.04624.

\bibitem{miyazaki24_interspeech}
K.~Miyazaki, Y.~Masuyama, and M.~Murata, ``{Exploring the Capability of Mamba in Speech Applications},'' in \emph{{Interspeech}}, Kos, Greece, 2024.

\end{thebibliography}

\vspace{12pt}
\color{red}

\end{document}